%% file: neurips_2023.tex
\documentclass{article}

\usepackage[preprint]{neurips_2023}

\usepackage[utf8]{inputenc} 
\usepackage[T1]{fontenc}    
\usepackage{hyperref}       
\usepackage{url}            
\usepackage{booktabs}       
\usepackage{amsfonts}       
\usepackage{nicefrac}       
\usepackage{microtype}      
\usepackage{xcolor}         

\definecolor{myblue}{RGB}{77,170,247}
\definecolor{myyellow}{RGB}{255,161,74}
\definecolor{mygreen}{RGB}{121,231,166}
\usepackage{graphicx}
\usepackage{amsmath}
\usepackage{amssymb}
\usepackage{multirow}
\usepackage{algorithm}
\usepackage{algorithmic}

\newcommand{\ie}{\textit{i.e.}}

\title{NEUCORE: Neural Concept Reasoning for Composed Image Retrieval}

\author{%
  Shu Zhao \\
  Pennsylvania State University\\
  University Park, USA\\
  \texttt{smz5505@psu.edu} \\
  \And
  Huijuan Xu\\
  Pennsylvania State University\\
  University Park, USA\\
  \texttt{hkx5063@psu.edu} \\
}

\begin{document}

\maketitle

\input{sections/0_abstract.tex}
\input{sections/1_introduction.tex}
\input{sections/2_related_work.tex}
\input{sections/3_method.tex}
\input{sections/4_experiments.tex}
\input{sections/5_conclusion.tex}

\bibliographystyle{plainnat}
\bibliography{reference}

\newpage
\appendix
\input{sections/appendix}

\end{document}

%% file: sections/0_abstract.tex
\begin{abstract}
  Composed image retrieval which combines a reference image and a text modifier to identify the desired target image is a challenging task, and requires the model to comprehend both vision and language modalities and their interactions. Existing approaches focus on holistic multi-modal interaction modeling, and ignore the composed and complimentary property between the reference image and text modifier. In order to better utilize the complementarity of multi-modal inputs for effective information fusion and retrieval, we move the multi-modal understanding to fine-granularity at concept-level, and learn the multi-modal concept alignment to identify the visual location in reference or target images corresponding to text modifier. Toward the end, we propose a \textbf{NEU}ral \textbf{CO}ncept \textbf{RE}asoning~(NEUCORE) model which incorporates multi-modal concept alignment and progressive multi-modal fusion over aligned concepts. Specifically, considering that text modifier may refer to semantic concepts not existing in the reference image and requiring to be added into the target image, we learn the multi-modal concept alignment between the text modifier and the concatenation of reference and target images, under multiple-instance learning framework with image and sentence level weak supervision. Furthermore, based on aligned concepts, to form discriminative fusion features of the input modalities for accurate target image retrieval, we propose a progressive fusion strategy with unified execution architecture instantiated by the attended language semantic concepts. Our proposed approach is evaluated on three datasets and achieves state-of-the-art results.
\end{abstract}

%% file: sections/1_introduction.tex
\section{Introduction}

\begin{figure}[t]
    \begin{center}
    \includegraphics[width=0.7\linewidth]{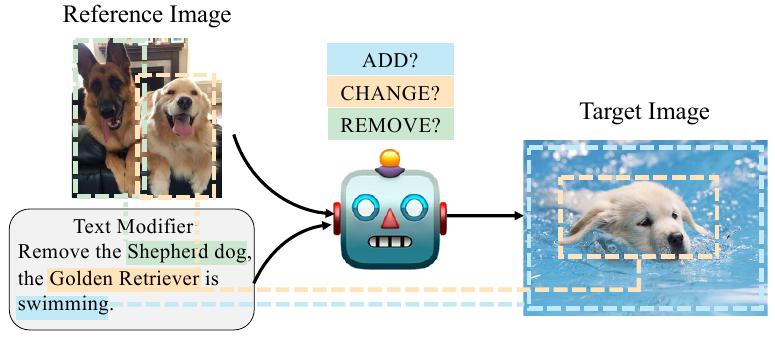}
    \end{center}
    \caption{Example of composed image retrieval. Visual concepts are mined from images and aligned with semantic concepts from the text modifier. Based on aligned concepts, the reference image feature is fused with the text modifier in a sequential way to identify the target image feature. Different colors denote different concepts. Note that a concept from text modifier may appear in the reference image~(\textcolor{mygreen}{Shepherd Dog}), the target image~(\textcolor{myblue}{Swim}), or both~(\textcolor{myyellow}{Golden Retriever}).}
    \label{fig:pro}
\end{figure}

Composed image retrieval~\citep{TIRG,VAL,CIRR,ARTEMIS} aims to identify target image, corresponding to the input query composed of a reference image and a text modifier describing how the reference image should be modified, as illustrated in Figure~\ref{fig:pro}. Compared to traditional image-to-image retrieval task~\citep{reid1,CCDH,RDH,img2img} and text-to-image retrieval task~\citep{txt2img1,txt2img2} where single modality is provided as input, composed image retrieval is a challenging task as it requires joint vision and language understanding to retrieve the corresponding target image.

Existing works tackle this problem by directly fusing the multi-modal features after single modality encoding~\citep{CIRR,DBLP:conf/cvpr/BaldratiBUB22a}. This type of approaches first process input modality as a whole and lack fine-grained multi-modal understanding at concept-level, preventing the model from performing concept-level composition, while most of the time the text modifier specifies partial semantic editing for the reference image. Therefore, we propose a \textbf{NEU}ral \textbf{CO}ncept \textbf{RE}asoning~(NEUCORE) model to mine and align the visual concepts in reference and target images with the semantic concepts in text modifier, to enhance the semantic consistency between the multi-modal feature composition and the target image feature. Specifically, our NEUCORE model consists of two main components, i.e., multi-modal concept alignment and progressive multi-modal fusion over concepts. 

Firstly, we propose to mine and align visual and semantic concepts under the weak supervision of image-text pair where the ground truth mapping of visual concepts in images and semantic concepts in sentences is unknown, instead of relying on object detectors~\citep{FasterRCNN} to generate region proposals and object tags to realize multi-modal concept alignment which suffers from the problems of heavy computational load and limited concept label space~\citep{VITCAP}. A multiple-instance learning framework with candidate visual concepts as instance, is designed with practical considerations of concept existence uncertainty, imbalance and noisy optimization, for tackling image-text pair weak supervision. Notably, a text modifier specifies the change to the reference image, and the change operations not only include attribute editing over the existing concepts in reference image, but also involve adding concepts into reference image or removing concepts from the reference image, which will cause the concept existence uncertainty among reference and target images. For example, in Figure~\ref{fig:pro}, the shepherd dog in the text modifier appears in the reference image, but the golden retriever appears in both the reference and target images. To overcome this uncertainty, we concatenate the reference and target image tokens corresponding to candidate visual concepts, and employ a transformer~\citep{Transformer} to jointly encode the reference and target images considering the nice property of patch-level feature encoding in transformer.

On top of visual encoding of reference and target images using transformer, the optimization for visual and semantic concept alignment is achieved by attention based multiple instance learning~\citep{DBLP:conf/icml/IlseTW18} under the supervision of the semantic concepts parsed from the text modifier. Practically, the semantic concepts from each text modifier are represented by the multi-label representation in the concept label space which is constructed by the semantic concepts from all text modifiers in training data. Considering that each text modifier only mentions very limited semantic concepts in each pair of reference and target images compared to the large concept label space, there exists the problem of positive-negative imbalance during the optimization. Besides, the partial semantic editing property of the text modifier for the reference image may cause the situation that, some visual concepts are not referred by the text modifier and mislabeled as negative labels in the multi-label representation, according to the labeling rule that only the semantic concepts mentioned in the text modifier will be set as positive in the concept label space for that example. An asymmetric loss is applied to alleviate the imbalance and mislabeling problems during the concept alignment optimization. After the optimization, the visual tokens are assigned with semantic meaning and aligned with the semantic concepts referred in the text modifier.

Secondly, after aligning visual and semantic concepts, we are able to fuse multi-modal features at fine granularity instead of holistic visual and text features~\citep{ARTEMIS,CIRR} for final target image retrieval. A progressive multi-modal fusion module is proposed to gradually fuse the aligned concepts from the reference image and the text modifier in a sequential way with each step having its own focus. Progressive multi-modal fusion over concepts involves two sub-problems of how to generate the fusion operation sequence, and the specific design for each fusion operation. We propose a unified fusion operation design which can be instantiated by different sequence indicators to realize different fusion operations, through leveraging the advantage that the normalization layer can fuse its own preserved features, obtained from sequence indicators, with input visual features~\citep{DBLP:conf/cvpr/UlyanovVL17}. The unified design overcomes the time-consuming drawback of previous hand-crafted fusion designs and removes the dependence on expert knowledge~\citep{NMN,NSCL,NS-VQA}. The fusion operation sequence is sequentially generated by the co-attention of global encoding over individual word embedding in the text modifier, and each sequence step focuses on local semantic concept feature guided by the global semantic context of text modifier. The attended local semantic concept feature is the sequence indicator at each sequence step, and is used to drive the instantiation of the unified fusion operation, taking on the role of a meta-learner. Our proposed fusion sequence generation method is optimized with the final retrieval loss without single step supervision needed as in sequence-to-sentence methods~\citep{N2NMN,DBLP:conf/eccv/ChenLWW20,NSCL,NS-VQA,IEP}, and is able to deal with diverse sentences compared to language parser based approaches~\citep{NMN,DBLP:conf/naacl/AndreasRDK16}.

To summarize our contributions, we introduce a model NEUCORE for composed image retrieval consisting of a multi-modal concept alignment module and a multi-modal fusion module over aligned concepts. Our NEUCORE model learns fine-grained multi-modal concept alignment under image and sentence level weak supervision with actual influencing factors considered. Reference image and text modifier are progressively fused using a unified fusion operation over aligned concepts and under the sequence guidance of attended local semantic concepts, to gain a representation for target image retrieval with more semantic consistency.
We validate our proposed model on three datasets. The results show that our method consistently outperforms the state-of-the-art, demonstrating the effectiveness of our approach.

%% file: sections/2_related_work.tex
\section{Related Work}

\textbf{Composed Image Retrieval.} Composed image retrieval task receives a reference image and a text modifier describing how the reference image should be modified to obtain the desired target image. TIRG~\citep{TIRG} fuses the query image and text into one multi-modal feature vector through a gating and residual mechanism. VAL~\citep{VAL} designs a composite module inserted in multiple layers of the visual encoder to preserve the visual information and modify it according to the text modifier. DCNet~\citep{DCNet} devises a Correction Network as a regularizer to maintain semantic consistency before and after the feature composition process for better text modification. CoSMo~\citep{CoSMo} disentangles features into content and style by introducing a content modulator and a style modulator, and then composes them with text modifiers to retrieve target images conditioned on both content and style information. CIRPLANT~\citep{CIRR} employs a large pre-trained vision-language model named OSCAR~\citep{OSCAR} to fuse vision and language information by leveraging the rich knowledge in the pre-trained model. ARTEMIS~\citep{ARTEMIS} achieves impressive results by decomposing the task into an image-image retrieval task and a text-image retrieval task, employing two attention modules for image-image and text-image, and then fusing these attended features to retrieve target images. 
However, these previous methods encode the reference image or text modifier into a holistic feature representation, and ignore the fine-grained information  in composed image retrieval where the composed and complimentary property between the reference image  and text modifier inputs happens. In this paper, our approach moves the multi-modal understanding to fine granularity at concept-level, and models the interactions between visual and semantic concepts for composed image retrieval.

\textbf{Multi-Modal Concept Alignment.} Multi-modal concept alignment aims to align visual concept space with semantic concept space to enable the fine-grained understanding and interaction between visual and semantic concepts~\citep{DBLP:conf/eccv/LiH022,DBLP:conf/naacl/LuoGGKD022}. A line of methods~\citep{KD-VLP,DBLP:conf/naacl/LiYWZCC21,DBLP:conf/aaai/ZhouPZHCG20,OSCAR,LXMERT} employ a pre-trained object detector~\citep{FasterRCNN} to generate region proposals and their object tags as the visual and semantic concepts, and then align them for downstream tasks. However, the object detectors limit the number of concepts as they are typically trained on limited pre-defined object categories. Some works investigate detector-free-based methods that use grid~(patch) features~\citep{VITCAP,DBLP:conf/cvpr/JiangMRLC20,RelVit} to predict visual concepts under weak supervision. In this paper, our model mines and aligns multi-modal concepts under the setting that the text modifier only contains partial semantic editing property, which is challenging to find correspondence to align concepts for composed image retrieval.

\textbf{Multi-Modal Fusion.} Multi-modal fusion approaches blend visual and text features for various vision and language tasks. Two categories of methods are typically used for multi-modal fusion. One is concatenating the visual and text tokens together and feeding them into a transformer to get fused features~\citep{DBLP:conf/aaai/ZhouPZHCG20,DBLP:conf/cvpr/DouXGWWWZZYP0022}. The second category of methods is employing cross-attention to exchange and fuse information between visual and text tokens~\citep{LXMERT,DBLP:conf/nips/LiSGJXH21}.

Moving forward, to improve the semantic consistency for target image retrieval in composed image retrieval, our method fuses the visual and text features progressively in a sequential way on top of the aligned fine-grained multi-modal concepts. To generate a fusion sequence, some works employ language parsers to extract the sequence from language modality~\citep{NMN,DBLP:conf/naacl/AndreasRDK16}. However, language parsers may not perform well on diverse sentences and lack generalization ability. A second typical method is to learn the sequence from data by training a sequence-to-sequence model to decode the action sequence from the input text~\citep{N2NMN,DBLP:conf/eccv/ChenLWW20,NSCL,NS-VQA,IEP}, while these methods typically need sequence level annotation which are annotation expensive. Our method learns the fusion sequence through global query driven co-attention over local semantics without sequence level annotation and ensures the generalization ability, and each fusion step is a unified design with specific instantiation controlled by the meta-learner from the generated fusion sequence, so there is a focus at each step in the progressive fusion process to improve the semantic consistency for target retrieval. 

%% file: sections/3_method.tex
\section{Method}

\begin{figure*}[t]
    \begin{center}
    \includegraphics[width=0.9\linewidth]{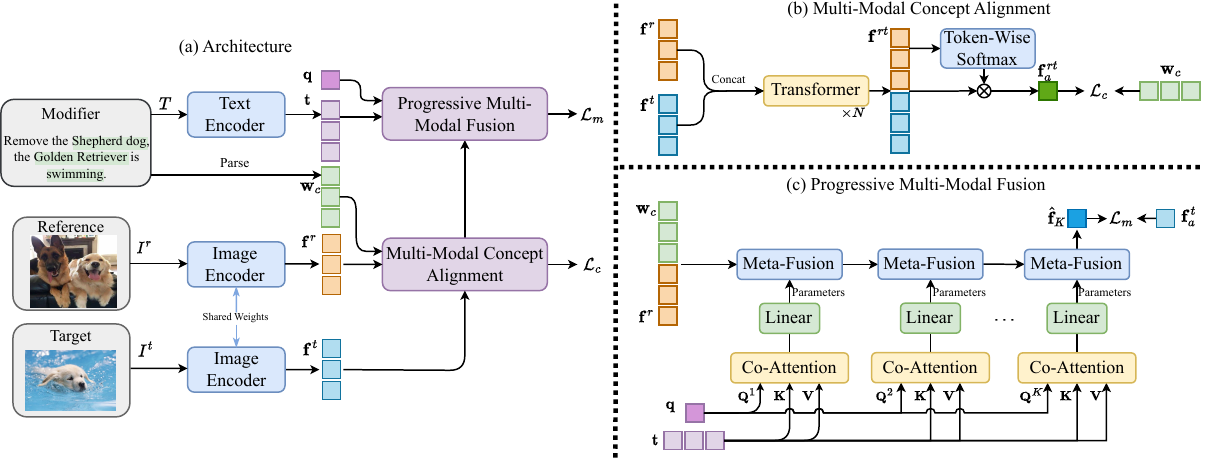}
    \end{center}
    \caption{(a) The overall architecture of our proposed NEUCORE model. It mines and aligns multi-modal concepts, then fuses the reference image feature with the text modifier over aligned concepts to identify the target image feature. (b) Multi-modal concept alignment module mines visual concepts from images under image-level supervision and aligns them with semantic concepts from text modifiers. (c) Progressive multi-modal fusion module decomposes the text modifier to a fusion sequence by co-attention and progressive fuses the reference image and text modifier over aligned concepts.}
    \label{fig:arch}
\end{figure*}

Given a reference image $I^r$ and a text modifier $T$, the composed image retrieval task aims to combine them to identify the target image $I^t$. Previous approaches holistically process each input modality and then fusion, and is lack of the fine-grained compositional understanding. In this paper, we propose Neural Concept Reasoning~(NEUCORE) to tackle the composed image retrieval by mining and aligning multi-modal concepts, and progressively fusing input modalities over concepts, as illustrated in Figure~\ref{fig:arch}. The explanation of the symbols used in this paper is listed in supplementary material.

For feature encoding, the text modifier feature $\mathbf{q}$ and contextualized word features $\mathbf{t}$ are encoded by a text encoder $\operatorname{E}_T$. An image encoder $\operatorname{E}_I$ is employed to extract reference and target image features, and the reference tokens $\mathbf{f}^{r}$ and target tokens $\mathbf{f}^{t}$ are obtained by flattening encoded visual features:
\begin{equation} \label{eqn:vis_enc}
    \begin{aligned}
        \mathbf{q}, \mathbf{t} = \operatorname{E}_{T}(T), \mathbf{f}^{r} = \operatorname{E}_{I}(I^r), \mathbf{f}^{t} = \operatorname{E}_{I}(I^t),
    \end{aligned}
\end{equation}
where superscripts $^{r}$ and $^{t}$ indicate that a feature belongs to the reference or target image, respectively.

\subsection{Multi-Modal Concept Alignment}
To model the fine-grained vision and language alignment between reference image and text modifier, we propose to mine and align visual concepts with semantic concepts from image-text pair data, which surpasses previous detector-based methods limited to small pre-defined label space. Due to the lack of concept-level supervision, we extract semantic concepts from text modifiers using a language parser as pseudo labels and apply the pseudo labels as image-level supervision. However, it is still challenging to learn multi-modal concept alignment with pseudo semantic concept labels at image level. Specifically, given a semantic concept, we cannot determine whether the correspondent visual concept appears in reference image or target image. For example, a text modifier ``\texttt{Remove a dog}'' means a \texttt{dog} concept exists in the reference image, not in the target image. And a text modifier ``\texttt{Add a cat}'' denotes a cat concept belonging to the target image, not in the reference image. Formally, given an input~($I^r$, $I^t$, or $T$), a concept set contains all concepts in the input, denoted as $C(\cdot)$. However, we cannot determine $c \in C(I^r)$ or $c \in C(I^t)$, where $c \in C(T)$. To resolve this ambiguity, we combine visual tokens of $I^r$ and $I^t$ to form a larger token set $I^{rt} = [I^r, I^t]$ and $C(T)\subset C(I^{rt})$, where $[\cdot,\cdot]$ denotes concatenation, considering that a concept $c$ described in $T$ must exist in $I^r$ or $I^t$~(or both).

Specifically, we parse\footnote{https://spacy.io/} semantic concepts from the text modifier and embed each semantic concept $\mathbf{w}_c$ via GloVe~\citep{GloVe} word embedding:
\begin{equation} \label{eqn:emb_concept}
    \begin{aligned}
        \mathbf{w}_c &= \operatorname{Embedding}(c).
    \end{aligned}
  \end{equation}
where $c \in \mathbb{M}$ and $\mathbb{M}$ is the concept vocabulary constructed by semantic concepts from all text modifiers.

Then, we obtain $\mathbf{f}^{rt}$ by concatenating reference and target tokens, and employ a transformer $\operatorname{Trans}$ to exchange context and find correspondence:
\begin{equation} \label{eqn:transformer}
    \begin{aligned}
       \mathbf{f}^{rt} &= \operatorname{Trans}([\mathbf{f}^{r}, \mathbf{f}^{t}]).
    \end{aligned}
\end{equation}

After modeling relation between reference and target tokens, we adopt a token-wise softmax to acquire attention weights $\mathbf{a}$, and then use the weights to summarize visual tokens to get a visual concept feature $\mathbf{f}_{a}^{rt}$ which contains visual foreground information:
\begin{equation} \label{eqn:att_head}
    \begin{aligned}
        \mathbf{a} = \operatorname{Softmax}(\mathbf{f}^{rt}), \mathbf{f}_{a}^{rt} = \mathbf{a} \cdot \mathbf{f}^{rt}.
    \end{aligned}
\end{equation}

Finally, multi-modal alignment score $\mathbf{s}$ is calculated by the concept features and embeddings of semantic concepts:
\begin{equation} \label{eqn:align_score}
  \begin{aligned}
      \mathbf{s} = \mathbf{f}_{a}^{rt} \cdot \mathbf{w}_c.
  \end{aligned}
\end{equation}

However, employing vanilla classification loss functions, such as the binary cross-entropy loss function, to optimize the score $\mathbf{s}$ leads to poor alignment. Text modifier only explicitly describes partial concepts compared to abundant visual concepts in the pair of input images, denoted as $\overline{C}(I^{rt}) = C(I^{rt}) - C(T)$ and $C(T) \subsetneqq C(I^{rt})$. However, all concepts belonging to $\overline{C}(I^{rt})$ are viewed as negative labels, leading to an increased risk of misclassifying visual concepts as background. On the other hand, the number of concept categories $\vert \mathbb{M}\vert$ is much larger than positive concepts described in each text modifier, \ie, $\vert C(I^{rt}) \vert \ll \vert \mathbb{M} -  C(I^{rt}) \vert$, causing high imbalance between positive and negative concept labels in the multi-label representation. As a result, the problems of mislabeling and imbalance hurt the training process, leading to incorrect concept alignment.

Therefore, we introduce an asymmetric loss~\citep{ASL} for multi-modal concept alignment to balance the visual and semantic concepts dynamically and discard possibly mislabeled concepts:
\begin{equation}\label{eqn:asl}
    \begin{aligned}
        \mathbf{s}^{'} &= \operatorname{sigmoid}(\mathbf{s}) \\
        \mathcal{L}_{\text{c}} &= -\frac{1}{N}\left(\sum_{i \in \mathbb{P}}(1 - \mathbf{s}_i^{'})^{\beta_+}\log(\mathbf{s}_i^{'})+\sum_{j \in \mathbb{N}}(\mathbf{s}_j^{'})^{\beta_-}\log(1 - \mathbf{s}_j^{'})\right),
    \end{aligned}
\end{equation}
where $\mathbb{P}$ and $\mathbb{N}$ are the positive and negative set, respectively. $\beta+$ and $\beta-$ are two hyper-parameters that control the degree of focus on positive and negative concepts.

\subsection{Progressive Multi-Modal Fusion over Concepts}
After obtaining aligned multi-modal concepts, the reference image feature $\mathbf{f}^{r}$ and text modifier feature $\mathbf{q}$ are progressively fused in a sequential way over aligned concepts to identify the target concept feature $\mathbf{f}^{t}_{a}$. We propose to decompose the sequential fusion steps from the text modifier without step level supervision. Specifically, to focus on distinct semantic contexts of a text modifier, $K$ independent fully connected layers $\operatorname{FC}_i, i=1,2,\cdots,K$ are employed to project text modifier feature $\mathbf{q}$. We adopt the Multi-Head Attention~($\operatorname{MHA}$)~\citep{Transformer} to extract the indicator vector for each semantic fusion step in the fusion sequence $\mathbb{S}$:
\begin{equation}\label{eqn:fs}
    \begin{aligned}
        \mathbb{S} = (\operatorname{MHA}(\operatorname{FC}_1(\mathbf{q}), \mathbf{t}, \mathbf{t}), \operatorname{MHA}(\operatorname{FC}_2(\mathbf{q}), \mathbf{t}, \mathbf{t}), \cdots, \operatorname{MHA}(\operatorname{FC}_K(\mathbf{q}), \mathbf{t}, \mathbf{t})) 
    \end{aligned}
\end{equation}

To progressively combine the reference image and text modifier over aligned concepts, we propose to instantiate specific operators from a meta-fusion architecture according to the generated fusion steps, which surpasses previous methods with time-consuming hand-crafted fusion operator design and the limit to expert knowledge. Our basic idea is that fusion steps can be clustered into semantic fusion groups, like ``\texttt{ADD}'' and ``\texttt{REMOVE},'' although the expressions within each group may vary in natural language. Therefore, we devise a transformer-based~\citep{Transformer} meta-fusion module~$\operatorname{MetaFusion}$ and allow it to be instantiated for specific semantic fusion groups. Specifically, we employ a fully connected layer to generate parameters $\operatorname{FC}(\mathbf{S}_i)$ according to fusion steps' indicators, where $i=1,2,\cdots,K$ and $\mathbf{S}_i \in \mathbb{S}$, and initialize the normalization layers~\citep{AdaIN} in the transformer with these parameters: 
\begin{equation} \label{eqn:ins_gen}
   \begin{aligned}
        \mathbf{\mu}_i = \operatorname{FC}(\mathbf{S}_i),\\
        \mathbf{\sigma}_i = \operatorname{FC}(\mathbf{S}_i).
   \end{aligned} 
\end{equation}

After transformer fusion instantiation, reference image tokens exchange information with aligned concepts in multi-head attention and fuse with text modifier information in the normalization layer:
\begin{equation} \label{eqn:fusion_step}
    \begin{aligned}
        \hat{\mathbf{f}}^{'}_{i-1} &= \operatorname{NL}(\hat{\mathbf{f}}_{i-1}; \mathbf{\mu}_i, \mathbf{\sigma}_i),\\
        \mathbf{Q}, \mathbf{K}, \mathbf{V} &= \mathbf{W} \hat{\mathbf{f}}^{'}_{i-1},\\
        \hat{\mathbf{f}}^{''}_{i-1} &= \operatorname{MHA}(\mathbf{Q}, \mathbf{K}, \mathbf{V}) + \hat{\mathbf{f}}^{'}_{i-1},\\
        \hat{\mathbf{f}}^{'''}_{i-1} &= \operatorname{NL}(\hat{\mathbf{f}}^{''}_{i-1}; \mathbf{\mu}_i, \mathbf{\sigma}_i),\\
        \hat{\mathbf{f}}_i &= \operatorname{FFN}(\hat{\mathbf{f}}^{'''}_{i-1}) + \hat{\mathbf{f}}^{'''}_{i-1},
    \end{aligned}
\end{equation}
where $\operatorname{NL}$ is a normalization layer; $\operatorname{FFN}$ is a feedforward network; $\mathbf{W}$ is a weight matrix; $\hat{\mathbf{f}}_{0}=\mathbf{f}^{r}$. After $K$ fusion steps, we obtain $\hat{\mathbf{f}}_K$ as the modified image feature to match the target images.

Finally, the concept matching score $\mathbf{m}$ can be calculated by:
\begin{equation} \label{eqn:matching}
    \begin{aligned}
        \mathbf{m} = \hat{\mathbf{f}}_K \cdot \mathbf{f}^t_{a},
    \end{aligned}
\end{equation}
where $\mathbf{f}^t_{a}$ is the visual concept feature of target image.

To train our proposed NEUCORE model, given mini-batch data, the model is optimized by the batch-based classification loss, which is demonstrated to be an efficient optimization loss function for the composed image retrieval task in previous approaches~\citep{TIRG,CoSMo,ARTEMIS}:
\begin{equation} \label{eqn:loss_sim}
    \begin{aligned}
        &\mathcal{L}_{m} = -\frac{1}{N}\sum_{i=1}^{N}\log\frac{\exp\{\gamma\mathbf{m}(\mathbf{I}^r_i, \mathbf{T}_i, C(I^r_i), \mathbf{I}^t_i)\}}{\sum_{j}\exp\{\gamma\mathbf{m}(\mathbf{I}^r_i, \mathbf{T}_i, C(I^r_i), \mathbf{I}^t_j)\}},\\
    \end{aligned}
\end{equation}
where $\gamma$ is a temperature parameter. $\mathbf{m}(\mathbf{I}_r^i, \mathbf{T}^i, C(I_r^i), \mathbf{I}_t^i)$ is the matching score~(cosine similarity), consisting of the concept matching score which is described in Equation~\eqref{eqn:matching} and the context matching score which is described in~\citep{ARTEMIS}. Combining with the loss function of concept alignment, the final loss function is:
\begin{equation} \label{eqn:loss}
    \begin{aligned}
        \mathcal{L} = \mathcal{L}_m + \alpha\mathcal{L}_c,
    \end{aligned}
\end{equation}
where $\alpha$ control the trade-off between the two loss functions.

The learning algorithm of our NEUCORE model is summarized in supplementary material. 

%% file: sections/4_experiments.tex
\section{Experiments}

\subsection{Datasets}

To evaluate our NEUCORE model, we extensively conduct experiments on three widely used datasets: Shoes~\citep{shoes}, FashionIQ\citep{FashionIQ}, and CIRR~\citep{CIRR}.

\textbf{Shoes} is constructed from the Attribute Discovery Dataset~\citep{ADD}. In~\citep{shoes}, authors additionally label natural language query sentences for the composed image retrieval task based on attribute labels in the original dataset. The dataset consists of 9k training triplets and 1.7k test queries.

\textbf{FashionIQ} covers three fashion categories: Dress, Top tee, and Shirt. It contains 46k images for training, and 15k images for validation and testing. There are 18k queries for training, 12k queries for validation, and 12k queries for testing. Each query has two captions describing how to modify from the reference image to the target image.

\textbf{CIRR} consists of over 36k open-domain images with human-generated text modifier. It is a more challenging dataset due to the richness of visual information and the diversity of language queries. Following~\citep{CIRR}, 36k triplets are split into 80\% for training, 10\% for validation, and 10\% for testing in our experiments.

\subsection{Evaluation Protocol}

Following~\citep{ARTEMIS}, we report composed image retrieval performance in Recall within top-$K$~($R@K$). Particularly, for CIRR dataset, following~\citep{CIRR}, we additionally report Recall within top-$K$ on the visually similar subset~($R_{s}@K$), where the subset of candidate target images is visually similar to the correct target image, and it requires the fine-grained understanding ability of both vision and language modalities and their interactions. Following previous works~\citep{ARTEMIS}, we evaluate NEUCORE model on the test set of CIRR dataset, the validation set of Shoes dataset, and the validation set of FashionIQ dataset.

\subsection{Implementation Details}

We employ ResNet~\citep{ResNet} pre-trained on ImageNet as the image encoder. For the text encoder, we adopt BiGRU~\citep{BiGRU} to encode sentence $\mathbf{q}$ and obtain the hidden states as contextualized word features $\mathbf{t}$. The concept alignment module consists of 2 transformer layers~\citep{ViT}. The batch size is 32. Following ~\citep{ARTEMIS}, we freeze the image encoder during the first 8 epochs. Then the model is trained for 50 epochs. We use the AdamW optimizer~\citep{AdamW} and set the initial learning rate to $5 \times 10^{-4}$ with a decay of 0.5 every 10 epochs. $\beta_{+}$ and $\beta_{-}$ are 1 and 4, respectively. $K$ is 3. $\gamma$ is $2.65926$. $\alpha$ is 200. The model is trained on one NVIDIA RTX A5000 GPU.

\subsection{Main Results}
We compare the performance of our proposed NEUCORE model with previous SOTA works on three benchmarks.

\input{tables/tab_cirr}
\textbf{Results on CIRR Dataset.} Table~\ref{tab:CIRR} shows the results. Our proposed NEUCORE model achieves $3.79$ average gain compared to the SOTA approach, ARTEMIS. Specifically, NEUCORE model improves $1.5$, $3.3$, $2.26$, and $1.62$ in $K=1$, $K=2$, $K=10$, and $K=50$ of $R@K$, respectively. It demonstrates that NEUCORE model can better retrieve target images according to reference images and text modifiers. Moreover, NEUCORE model improves $4.28$, $4.86$, and $3.25$ in $K=1$, $K=2$, and $K=3$ of $R_{s}@K$, indicating the recall of top-$K$ on a visual similar subset, which is a more challenging metric and evaluates a model's fine-grained understanding ability on a visual similar target subset.
Under the same pre-training condition without using the vision-language pre-trained weights, denoted as CIRPLANT, our model outperforms CIRPLANT significantly on all metrics. Furthermore, when the CIRPLANT model is initialized with the pre-trained weights from the OSCAR model~\citep{OSCAR} trained on $6.5$ million image-caption pairs, our model can still outperform it in $R_s@1$ and $R_s@2$ by 5.07 and 4.03, respectively. It indicates our proposed multi-modal concept alignment module can effectively mine and align fine-trained multi-modal concepts and the multi-modal fusion can fuse the reference image feature and text modifier feature over concepts to identify the target image feature.

\input{tables/tab_shoes}
\textbf{Results on Shoes Dataset.} The results are shown in Table~\ref{tab:Shoes}. Our proposed NEUCORE model improves $1.04$, $2.37$, and $1.44$ in $K=1$, $K=10$, and $K=50$ of Recall compared to the SOTA method ARTEMIS~\citep{ARTEMIS} on this dataset. And it achieves $1.62$ improvement of average improvement, $(R@1+R@10+R@50)/3$.

\input{tables/tab_fashionIQ}
\textbf{Results on FashionIQ Dataset.} Table~\ref{tab:FashionIQ} illustrates the main results. Our proposed NEUCORE model still obtains state-of-the-art results of $R@10$ and $R@50$. Due to the limited space, detailed results with more evaluation metrics on the FashionIQ dataset can be found in supplementary material.

\begin{figure*}[!t]
    \begin{center}
    \includegraphics[width=0.8\linewidth]{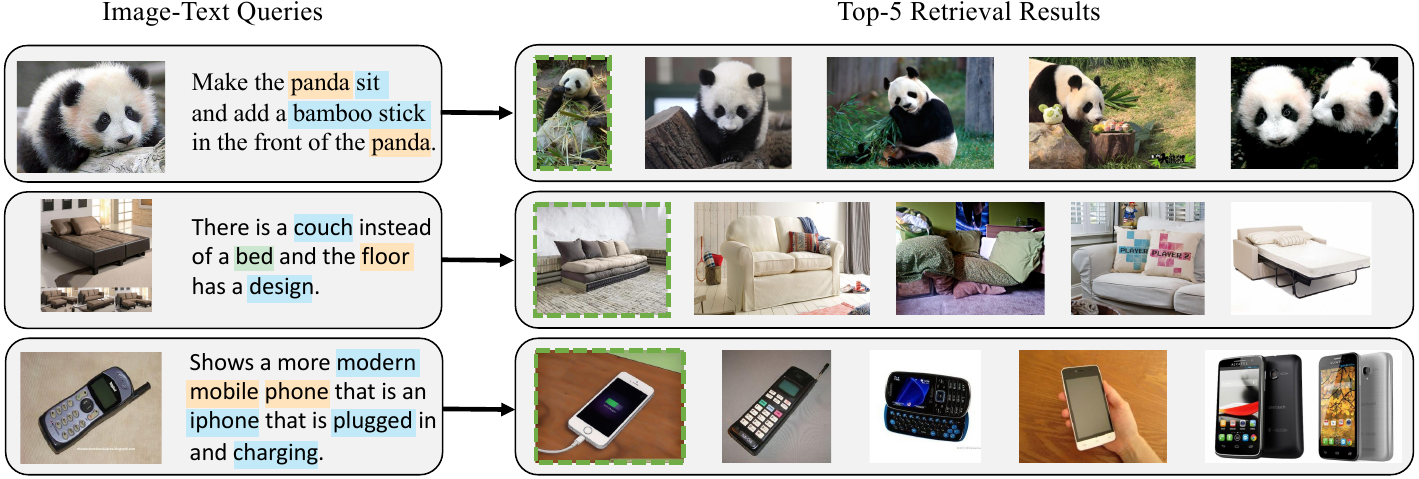}
    \end{center}
    \caption{Qualitative examples of image-text queries on CIRR validation set and its Top-5 retrieval results. Green dotted boxes denote the ground-truth target images, and semantic concepts in the input text modifiers are highlighted by different colors. \textcolor{mygreen}{Green} indicates that a concept appears only in the reference image. \textcolor{myyellow}{Yellow} denotes that a concept appears both in the reference and target images. \textcolor{myblue}{Blue} means that a concept appears only in the target image.}
    \label{fig:qualitative}
\end{figure*}

\subsection{Ablation Study}
To show the effectiveness of each component of our model design, ablation study for different variants are conducted on the CIRR validation set as CIRR dataset contains more concepts and is better to evaluate the fine-grained understanding ability of models.

\input{tables/tab_ablation_concept}
\textbf{Multi-modal Concept Alignment Module}. The ablation study results of the concept alignment module are illustrated in Table~\ref{tab:ablation_concept}. ``\textit{Reference Only}'' and ``\textit{Target Only}'' denote that we do not concatenate the reference and target image tokens and only use reference or target image tokens to align with semantic concepts. The performance decrease of these two variants demonstrates that our concept alignment module with the concatenation of reference and target images can help alleviate the ambiguity problem of corresponding visual concepts, and improve multi-modal concept mining and alignment for composed image retrieval task.
``\textit{Cross-Entropy Loss}'' means that the asymmetric loss for optimizing the concept alignment described in Equation~\eqref{eqn:asl} is replaced with the binary cross-entropy loss. The ablation results show the performance decrease, and confirm the existence of the problems about positive-negative imbalance and mislabeling in multi-label concept classification. Our proposed concept alignment module with asymmetric loss can help alleviate these problems.

\input{tables/tab_ablation_program}
\textbf{Progressive Multi-Modal Fusion Module}. The ablation study results of the progressive multi-modal fusion module are shown in Table~\ref{tab:ablation_program}. 
It confirms the effectiveness of our progressive multi-modal fusion module with automatic fusion sequence generation and unified fusion module design. ``\textit{Remove Progressive Fusion module}'' means that we remove the multi-modal progressive fusion module and directly fuse concatenation of the reference image feature and text modifier feature. The decreased ablation results demonstrate that our proposed fusion module can better identify the target feature by progressively fusing the reference image feature and text modifier feature over aligned concepts with each step having focus. ``\textit{Layer norm}'' denote we use vanilla layer normalization instead of adaptive instance normalization without adaptive instantiation. The results show adaptive instance normalization can fuse features better in composed image retrieval task.

\input{tables/tab_zero_shot}

\subsection{Analysis of Zero-Shot Concepts}
Compared with previous approaches, our proposed NEUCORE model mines and aligns visual concepts and semantic concepts. It aligns the visual embedding space and semantic embedding space and transfers knowledge from language to vision, which can improve the zero-shot concept recognition ability. To demonstrate it, we create a data split from CIRR validation set, named $\text{CIRR}_{\text{zs}}$. Specifically, we parse the concepts from the training and validation sets. Next, we compute their difference set to obtain zero-shot concepts, \ie, not seen during training time. Then we only keep these samples that contain zero-shot concepts to create zero-shot data split $\mathcal{D}_{\text{zs}}$, resulting in $316$ zero-shot concepts and $350$ samples.

Results are reported in Table~\ref{tab:zero_shot}. Compared to the SOTA approach ARTEMIS which fuses holistic multi-modal features for composed image retrieval, NEUCORE model improves $6.57$, $7.43$, and $4.28$ in $K=1$, $K=2$, and $K=3$ of recall within top-K of subset $R_s$, respectively. Moreover, we also present the results of removing the multi-modal concept alignment module, which resulted in performance drops of $5.72$, $9.43$, and $6.28$ in $R@k$, respectively. It illustrates the concept alignment module improves the zero-shot concept recognition ability by aligning visual concepts with semantic concepts represented by word embedding, which transfers the knowledge from word embedding to visual concepts.

\subsection{Qualitative Results}
Figure~\ref{fig:qualitative} shows the retrieval examples from a restricted subset of the CIRR validation set~\cite{CIRR}, where candidate target images are visually similar. It requires learning fine-grained vision and language features and their interactions. The results show our proposed NEUCORE model can understand the content of text modifier and compose the reference image feature and text modifier feature to identify the target image feature.

%% file: tables/tab_cirr.tex
\begin{table*}[t]
    \caption{\textbf{Results on CIRR dataset.} CIRPLANT$^\star$ employs a large pre-trained vision-language model.
    \label{tab:CIRR} }
    \centering
    \resizebox{\linewidth}{!}{%
    \begin{tabular}{@{}lcccccccc@{}}
    \toprule
    \multirow{2}{*}{Method} & \multicolumn{4}{c}{$R@K$} & \multicolumn{3}{c}{$R_{s}@K$} & \multirow{2}{*}{$\frac{(R@5 + R_{s}@1)}{2}$} \\ \cmidrule(lr){2-5} \cmidrule(lr){6-8}
    & $K=1$ & $K=5$ & $K=10$ & $K=50$ & $K=1$ & $K=2$ & $K=3$ & \\
    \midrule
    CIRPLANT$^\star$~(init. OSCAR)~\cite{CIRR} & 19.55 & 52.55 & 68.39 & 92.38 & 39.20 & 63.03 & 79.49 & 45.88 \\
    \midrule
    CIRPLANT~\cite{CIRR} & {15.18} & {43.36} & {60.48} & {87.64} & 33.81 & 56.99 & {75.40} & {38.59} \\
    TIRG~\cite{TIRG} & 10.01 & 38.31 & 54.59 & 84.69 & {37.36} & {59.31} & 72.51 & 37.84 \\
    ARTEMIS~\cite{ARTEMIS} & {16.96} & {46.10} & {61.31} & {87.73} & {39.99} & {62.20} & {75.67} & {43.05} \\
    \midrule
    NEUCORE & \textbf{18.46} & \textbf{49.40} & \textbf{63.57} & \textbf{89.35} & \textbf{44.27} & \textbf{67.06} & \textbf{78.92} & \textbf{46.84}\\
    \bottomrule
    \end{tabular}
    }
    \end{table*}

%% file: tables/tab_shoes.tex
\begin{table}[t]
    \small 
    \caption{\textbf{Results on Shoes dataset.}  
    \label{tab:Shoes}
    }
    \centering
    \begin{tabular}{@{}lcccc@{}}
    \toprule
    Method & $R@1$ & $R@10$ & $R@50$ & $\frac{R@1+R@10+R@50}{3}$\\ \midrule
    TIRG~\cite{TIRG} & 14.46 & 47.51 & 75.17 & 45.71 \\ 
    VAL~\cite{VAL}  & 16.49 & 49.12 & 73.53 & 46.38 \\ 
    CoSMo~\cite{CoSMo} & 16.72 & 48.36 & 75.64 & 46.91 \\
    ARTEMIS~\cite{ARTEMIS} & {18.72} & {53.11} & {79.31} &  {50.38} \\ %
    \midrule
    NEUCORE & \textbf{19.76} & \textbf{55.48} & \textbf{80.75} & \textbf{52.00}\\
    \bottomrule
    \end{tabular}
\end{table}

%% file: tables/tab_fashionIQ.tex
\begin{table}[t]
    \caption{\textbf{Results on Fashion IQ dataset.}
    }
    \label{tab:FashionIQ}
    \centering
    \begin{tabular}{@{}lccc@{}}
    \toprule
    Method & $R@10$ & $R@50$ & $\frac{R@10+R@50}{2}$ \\
    \midrule
    CIRPLANT~\cite{CIRR} & 14.82 & 35.52 & 25.17 \\
    TIRG~\cite{TIRG} & 23.17 & 47.48 & 35.32 \\
    CoSMo~\cite{CoSMo} & 19.87 & 42.65 & 31.26 \\
    ARTEMIS~\cite{ARTEMIS} & 26.05 & 50.29 & 38.17 \\
    \midrule
    NEUCORE & \textbf{26.45} & \textbf{51.75} & \textbf{39.15} \\
    \bottomrule
    \end{tabular}
    \end{table}

%% file: tables/tab_ablation_concept.tex
\begin{table}[t]
    \caption{\textbf{Ablation study} of multi-modal concept alignment module on CIRR validation set.
    \label{tab:ablation_concept}}
    \centering
    \begin{tabular}{@{}lccc@{}}
    \toprule
    Variants & $R@5$ & $R_s@1$ & $\frac{(R@5 + R_{s}@1)}{2}$\\
    \midrule
    NEUCORE & \textbf{51.10} & \textbf{45.35} & \textbf{48.22}\\
    \midrule
    Reference Only & 50.74 & 42.76 & 46.75\\
    Target Only & 49.77 & 42.48 & 46.12\\
    Cross-Entropy Loss & 47.52 & 42.71 & 45.11\\
    \bottomrule
    \end{tabular}
\end{table}

%% file: tables/tab_ablation_program.tex
\begin{table}[t]
    \caption{\textbf{Ablation study} of multi-modal fusion module on CIRR validation set.
    \label{tab:ablation_program}}
    \centering
    \begin{tabular}{@{}lccc@{}}
    \toprule
    Variants & $R@5$ & $R_s@1$ & $\frac{(R@5 + R_{s}@1)}{2}$\\
    \midrule
    NEUCORE & \textbf{51.10} & \textbf{45.35} & \textbf{48.22}\\
    \midrule
    Remove Progressive Fusion Module & 50.29 & 43.92 & 47.10 \\
    Layer Norm & 49.32 & 44.73 & 47.03 \\
    \bottomrule
    \end{tabular}
    \end{table}

%% file: tables/tab_zero_shot.tex
\begin{table}[t!]
    \caption{\textbf{Analysis of zero-shot novel concept generalization} on the new data split $\text{CIRR}_\text{zs}$ with novel concepts in test data.
    \label{tab:zero_shot}}
    \centering
    \begin{tabular}{@{}lccc@{}}
    \toprule
    Method & $K=1$ & $K=2$ & $K=3$\\
    \midrule
    ARTEMIS~\cite{ARTEMIS} & 30.86 & 55.71 & 68.29\\
    \midrule
    NEUCORE & \textbf{37.43} & \textbf{63.14} & \textbf{72.57}\\
    Remove Concept Module& 31.71 & 53.71 & 66.29\\
    \bottomrule
    \end{tabular}
\end{table}

%% file: sections/5_conclusion.tex
\section{Conclusion}
In this paper, we propose a model named NEUCORE to tackle the composed image retrieval task, which consists of multi-modal concept alignment module and progressive multi-modal fusion module. Multi-modal concept alignment module mines and aligns visual concepts from images with semantic concepts from text modifiers, and the progressive multi-modal fusion module progressively fuses the reference image feature with the text modifier feature over aligned concepts to identify the target image feature. Extensive experiments demonstrate our proposed NEUCORE model learns fine-grained multi-modal alignment and their interactions at concept-level from image-text paired data.

%% file: sections/appendix.tex
\begin{table}[h]
    \caption{The list of symbols and notations used in this paper. $\star \in \{r, t, rt\}$ denotes the reference, target, or concatenation of reference and target.
    \label{tab:symbols}}
    \centering
    \begin{tabular}{@{}cl@{}}
    \toprule
    Symbol & Description\\
    \midrule
    $I^{\star}$ & raw image\\
    $\mathbf{f}^{\star}$ & visual tokens \\
    T & text modifier \\
    $\mathbf{t}$ & contextualized word features \\
    $\mathbf{q}$ & sentence feature \\
    $\mathbf{a}$ & attention weights \\
    $\mathbf{f}^{\star}_{a}$ & visual concept feature \\
    $\mathbf{s}$ & multi-modal alignment score \\
    $c$ & concept\\
    $C(\cdot)$ & concept set\\
    $\mathbf{w}_c$ & concept embedding \\
    $\mathbb{M}$ & concept vocabulary \\
    $\mathbb{S}$ & fusion sequence \\
    $\hat{\mathbf{f}}$ & modified feature \\
    $\mathbf{m}$ & concept matching score\\
    \bottomrule
    \end{tabular}
    \end{table}
\section{List of Symbols}
The list of symbols and notations used in this paper is shown in Table~\ref{tab:symbols}.

\begin{table}[H]
    \caption{\textbf{Concept Source}. Pseudo concepts labels are extract by a language parser according to part-of-speech on CIRR dataset.
    \label{tab:pos}}
    \centering
    \begin{tabular}{@{}lccc@{}}
    \toprule
    PoS & $R@5$ & $R_s@1$ & $\frac{(R@5 + R_{s}@1)}{2}$\\
    \midrule
    Noun & 49.12 & 43.20 & 46.16\\
    Noun + Adj & 49.25 & 43.28 & 46.27\\
    Noun + Adj + Verb & 50.00 & 44.01 & 47.01\\
    Noun + Adj + Verb + Adv & \textbf{51.10} & \textbf{45.35} & \textbf{48.22}\\
    \bottomrule
    \end{tabular}
\end{table}
\section{Concept Source}
Pseudo concept labels are extracted according to part-of-speech. To evaluate the effectiveness of different concept types, we extract ``Noun,'' ``Adj,'', ``Verb,'' and ``Adv'' from the CIRR validation set and combine them as pseudo labels. The results are shown in Table~\ref{tab:pos}. It demonstrates ``Noun + Adj + Verb + Adv'' achieves the best result, indicating that the number of concepts may affect the performance. More concepts help the model learn richer features.

\section{Detailed Results on Fashion IQ dataset}
\begin{table*}[!t]
    \caption{\textbf{Detailed results on Fashion IQ dataset}.
    }
    \label{tab:FashionIQ_full}
    \centering
    \resizebox{\linewidth}{!}{%
    \begin{tabular}{@{}lccccccccc@{}}
    \toprule
    \multirow{2}{*}{Method} &  \multirow{2}{*}{$\frac{R@10+R@50}{2}$} & \multicolumn{4}{c}{$R@10$} & \multicolumn{4}{c}{$R@50$} \\ \cmidrule(lr){3-6} \cmidrule(lr){7-10}
     & & Dress & Shirt & Toptee & Mean & Dress & Shirt & Toptee & Mean \\
    \midrule
    ComposeAE~\citep{ComposeAE} & 20.60 & - & - & - & 11.80 & - & - & - & 29.40 \\
    TCIR~\citep{TCIR} & 29.51 & 19.33 & 14.47 & 19.73 & 17.84 & 43.52 & 35.47 & 44.56 & 41.18 \\
    CIRPLANT~\citep{CIRR} & 25.17 & 14.38 & 13.64 & 16.44 & 14.82 & 34.66 & 33.56 & 38.34 & 35.52 \\
    TIRG~\citep{TIRG}  &
    35.32 & 23.80 & 19.90 & 25.82 & 23.17 & 48.64 & 42.14 & 51.64 & 47.48 \\
    VAL~\citep{VAL} & 33.82 & 21.12 & 21.03 & 25.64 & 22.60 & 42.19 & 43.44 & 49.49 & 45.04 \\
    CoSMo~\citep{CoSMo} &  31.26 & 21.39 & 16.90 & 21.32 & 19.87 & 44.45 & 37.49 & 46.02 & 42.65 \\
    ARTEMIS~\citep{ARTEMIS} &
         {38.17} & \textbf{27.16} & {21.78} & {29.20} & {26.05} & {52.40} & 43.64 & {54.83} & {50.29} \\
    \midrule
    NEUCORE & \textbf{39.15} & 27.00 & \textbf{22.84} & \textbf{29.63} & \textbf{26.45} & \textbf{53.79} & \textbf{45.00} & \textbf{56.65} & \textbf{51.75}\\
    \bottomrule
    \end{tabular}
    }
\end{table*}
Table~\ref{tab:FashionIQ_full} illustrates the detailed results on Fashion IQ validation set. It demonstrates that our proposed NEUCORE model can outperform the SOTA method ARTEMIS in most of the metrics. This also validates that progressive fusion with aligned multi-modal concept alignment can improve the composed image retrieval task.

\section{Qualitative Results}
\begin{figure*}[!t]
    \begin{center}
    \includegraphics[width=0.9\linewidth]{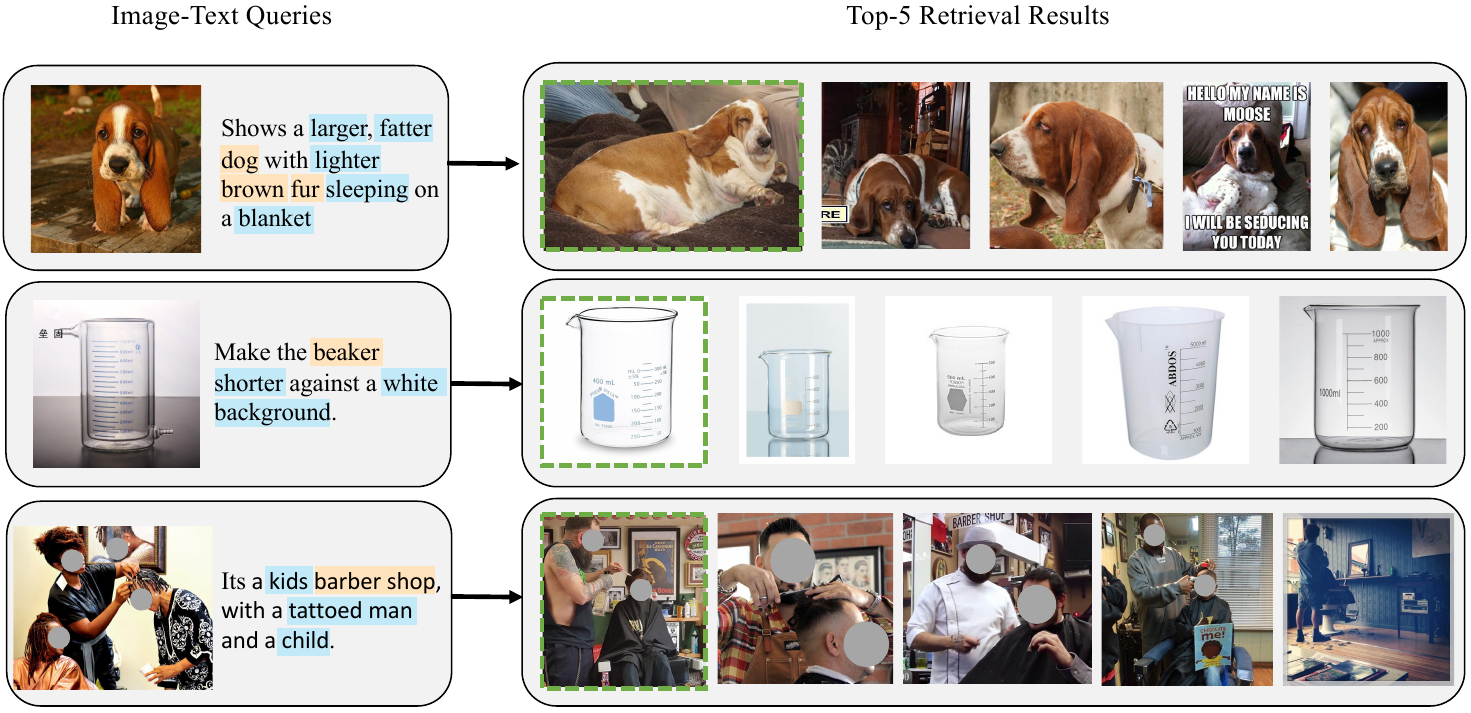}
    \end{center}
    \caption{Qualitative examples of image-text queries on CIRR validation set and its Top-5 retrieval results. Green dotted boxes denote the ground-truth target images, and semantic concepts in the input text modifiers are highlighted by different colors. \textcolor{myyellow}{Yellow} denotes that a concept appears both in the reference and target images. \textcolor{myblue}{Blue} means that a concept appears only in the target image.}
    \label{fig:appendix_qualitative}
\end{figure*}

We provide more qualitative retrieval examples from a restricted subset of the CIRR validation set~\citep{CIRR} where candidate target images are visually similar. It is challenging because the model needs to learn fine-grained vision and language features and their interactions. The retrieval examples are shown in Figure~\ref{fig:appendix_qualitative}. Results demonstrate our NEUCORE model can understand the content of text modifier, find correct correspondence between visual concepts and semantic concepts, and compose the reference image feature and text modifier feature to identify the target image feature.

\begin{figure*}[!t]
    \begin{center}
    \includegraphics[width=0.9\linewidth]{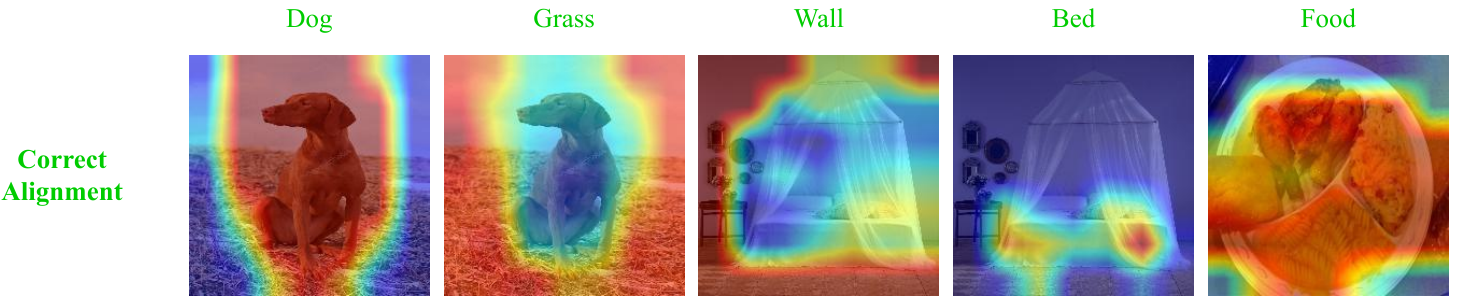}
    \end{center}
    \caption{The visualization results of correct concept alignment on CIRR dataset.}
    \label{fig:correct_alignment}
\end{figure*}

\begin{figure*}[!t]
    \begin{center}
    \includegraphics[width=0.7\linewidth]{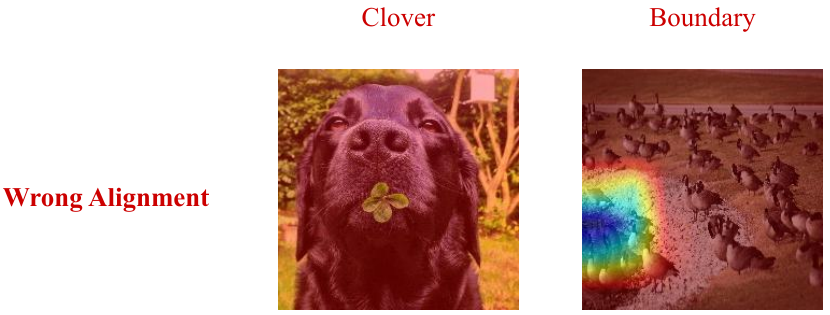}
    \end{center}
    \caption{The visualization results of wrong concept alignment on CIRR dataset.}
    \label{fig:wrong_alignment}
\end{figure*}

\section{Visualization of Concept Alignment}
Our NEUCORE model can mine and align visual concepts with semantic concepts. We employ Grad-CAM~\citep{gradcam} to identify visual concepts corresponding to semantic concepts on CIRR dataset. The correct visualization results are shown in Figure~\ref{fig:correct_alignment}. It demonstrates that our NEUCORE model can align semantic concepts to visual concepts in images under image level weak supervision.
We also show some failure visualization cases in Figure~\ref{fig:wrong_alignment}, where these semantic concepts are describing more abstract visual concepts. We expect that using more advanced vision and language pre-trained models can help alleviate these failure cases.

\section{List of Zero-shot Concepts on $\text{CIRR}\_{\text{zs}}$ Dataset}
To demonstrate our NEUCORE model can deal with novel zero-shot concepts, we create a data split from CIRR validation set, named $\text{CIRR}\_{\text{zs}}$. The zero-shot concepts in $\text{CIRR}\_{\text{zs}}$ are listed as follows:

plus, winglike, unicorn, lounger, camisole, wound, gentle, ibex, hinged, silicone, vary, jajantic, crayon, zone, servicing, rollaway, sorted, description, ended, secondhand, rop, softie, winner, makw, furnished, birn, moblie, hospital, orient, gose, bulk, cum, whote, pilot, hyppopotamus, sharo, simillar, thread, aniamls, celebration, committee, freestyle, scubadiver, decorated, counch, help, dia, fatter, goldtone, shovel, earing, huddled, florescent, handy, aggressive, sis, undecipherable, attitude, asembled, settle, celebrity, vietnamise, clapping, suv, cologne, wax, prepared, law, unrisen, boklane, screwlike, carcas, treet, monk, handleless, repaired, atheletic, biting, commercial, shrimp, entree, ticker, graze, intense, portait, buddy, swin, engineer, chemistry, unobvious, avenue, seashelle, gummie, misssig, empanada, puzzle, hazelnut, driverside, maze, seller, gnu, another, anmial, gummy, knob, mounted, thermal, furth, seagal, handing, wolflike, clap, barnlike, gape, clover, medusa, under, siringe, vegatation, raelistic, outrigger, cinnamon, brwon, screwtop, rocksurface, snorkeler, creative, wingspan, coastal, innocent, ostrich, volture, eatm, gril, cyclist, on, aless, pinecone, comedy, blueprint, spaghetti, inverted, fried, lea, treeline, unmanned, sweatshirt, swirly, silvertone, lengthy, coverge, attenae, multipack, stocking, hypopotamus, mosque, continental, baggage, stickering, core, silve, hollowed, swimwear, noodle, eyeliner, sphinx, multilayer, morevie, abdoman, hartebe, thumbnail, bouqet, boundary, glide, palican, boklaine, trianglular, steer, orthodox, assembly, canister, terrestrial, variable, spoonful, value, hyppopotomus, liner, wardrove, blackc, ribbontail, diameter, multimeter, sculture, droopy, hi, cachrro, removed, prayer, scent, italian, trimming, coulple, zest, calimari, jogger, frock, healthier, box(es, rabie, supine, sape, mkaing, dungeness, cubicle, last, unrenovated, miror, patterened, dool, pallet, femal, denser, mongoose, pelikan, mottled, mechanic, 50ml, affectionate, wintery, pitchet, cuttingboard, starbuck, thong, churchlike, handler, technology, hyypopotomus, bib, wipe, swollen, tong, hawaiian, lingerie, tupperware, hexagon, flatcap, omlette, makin, direciton, pearl, princess, hooded, façade, playin, skincare, smaler, screwdriver, spill, safe, throny, gopher, furier, embrace, luminous, choir, heard, taupe, bouse, unbrand, tentacled, carrier, gutted, oral, mouthed, squid, active, triney, landscaping, coffe, state, visibility, tortoise, buffet, erotic, bloody, barking, puppeis, account, risen, peper, anima, sandwedge, mold, guide, religious, largre, panty, sipper, differnt, chubbier, rwmove, welcoming, countryard, tattoed, compression, interacting, dove, veggie, descriptive, feminine, highway, orientation, bride, potote, disc, rocklike.

\section{Algorithm for Learning the NEUCORE model}
\begin{algorithm}[!t]
\caption{The overall learning algorithm for the NEUCORE model.}

\textbf{Input}: Reference image $I_r$; Target image $I_t$; Text modifier $T$.\\
\textbf{Output}: Matching score $\mathbf{m}$; Parameters of the NEUCORE model.

\begin{algorithmic}[1] 
\STATE Obtain semantic concepts $C(T)$ from a language parser.
\REPEAT
\STATE \textit{\# Encode vision and language features}
\STATE Extract reference visual tokens $\mathbf{f}^r$, target visual tokens $\mathbf{f}^t$, contextualized word features $\mathbf{t}$, and sentence feature $\mathbf{q}$ by Equation~(1).
\STATE
\STATE \textit{\# Multi-modal concept Alignment. Section 3.1}
\STATE Obtain semantic concept word embeddings $\mathbf{w}_c$ by Equation~(2).
\STATE Concatenate reference and target visual tokens to get concatenated tokens $[\mathbf{f}^{r}, \mathbf{f}^r]$ and feed them to transformer layer to exchange their information and obtain $\mathbf{f}^{rt}$ by Equation~(3).
\STATE Apply a token-wise softmax operation for concatenated tokens $\mathbf{f}^{rt}$ to get attention weights $\mathbf{a}$ and weighted summary concatenated tokens $\mathbf{f}^{rt}$ according to the attention weights $\mathbf{a}$ by Equation~(4).
\STATE Calculate the multi-modal alignment score $\mathbf{s}$ by Equation~(5).
\STATE 
\STATE \textit{\# Progressive multi-modal fusion over concepts. Section 3.2}
\STATE Generate fusion sequence $\mathbb{S}$ by Equation~(7).
\STATE Progressively fuse the reference image tokens $\mathbf{f}^r$ and the text modifier feature stored in fusion steps $\mathbb{S}$ over concepts $\mathbf{w}_c$ by Equation~(8).
\STATE Calculate the concept matching score $\mathbf{m}$ by Equation~(9).
\STATE 
\STATE \textit{\# Loss function}
\STATE Calculate the multi-modal concept alignment loss value $\mathcal{L}_c$ by Equation~(6).
\STATE Calculate the matching loss value $\mathcal{L}_m$ by Equation~(10).
\STATE Calculate the final loss value by Equation~(11) and optimize it by BP algorithm.
\UNTIL Convergence or reach maximum iterations.
\end{algorithmic}
\label{alg:training}
\end{algorithm}
The overall learning algorithm for our proposed NEUCORE model is illustrated in Algorithm~\ref{alg:training}.

\section{Limitations and Future Work}
Our proposed model, NEUCORE, can mine and align multi-modal concepts without concept-level supervision. However, the improvement in Fashion IQ is relatively small than CIRR. It is mainly because Fashion IQ contains domain-specific concepts, like ``suede'' and ``bely.'' CIRR is more diverse and covers more concepts. Recently, large vision-language (VL) models have achieved significant progress. Our model does not employ these VL models currently and only focuses on model side, but potentially these VL models could help our model learn more domain-specific concepts. On the other hand, we decompose the text modifier to generate a fusion sequence and progressively fuse the reference image feature and text modifier feature over aligned multi-modal concepts. Large language models~(LLM) can also be utilized to generate a more accurate fusion sequence.